\documentclass[10pt,twocolumn,letterpaper]{article}
\usepackage{cvpr}
\usepackage{colortbl}

\usepackage{float}
\definecolor{cvprblue}{rgb}{0.21,0.49,0.74}
\usepackage[pagebackref,breaklinks,colorlinks,allcolors=cvprblue]{h
yperref}
\usepackage{ragged2e}
\usepackage{booktabs,makecell, multirow, tabularx}
\def\paperID{1187} % *** Enter the Paper ID here
\def\confName{CVPR}
\def\confYear{2025}

\title{Scene4U: Hierarchical Layered 3D Scene Reconstruction from \\ Single Panoramic Image for Your Immerse Exploration}

\author{
Zilong Huang\textsuperscript{\rm 1} \quad Jun He\textsuperscript{\rm 1} \quad Junyan Ye\textsuperscript{\rm 1,2} \quad Lihan Jiang\textsuperscript{\rm 2, 3} \quad Weijia Li\textsuperscript{\rm 1} \quad Yiping Chen\textsuperscript{\rm 1 †} \quad Ting Han\textsuperscript{\rm 1 †} 
    \vspace{5pt} \\
    \textsuperscript{\rm 1} Sun Yat-sen University, 
    \textsuperscript{\rm 2} Shanghai Artificial Intelligence Laboratory, \\
    \textsuperscript{\rm 3} University of Science and Technology of China \\
}
\begin{document}

\twocolumn
[{%
\renewcommand\twocolumn[1][]{#1}%
\maketitle
\begin{center}
\vspace{-0.5cm}
\captionsetup{type=figure}
\includegraphics[width=\linewidth]{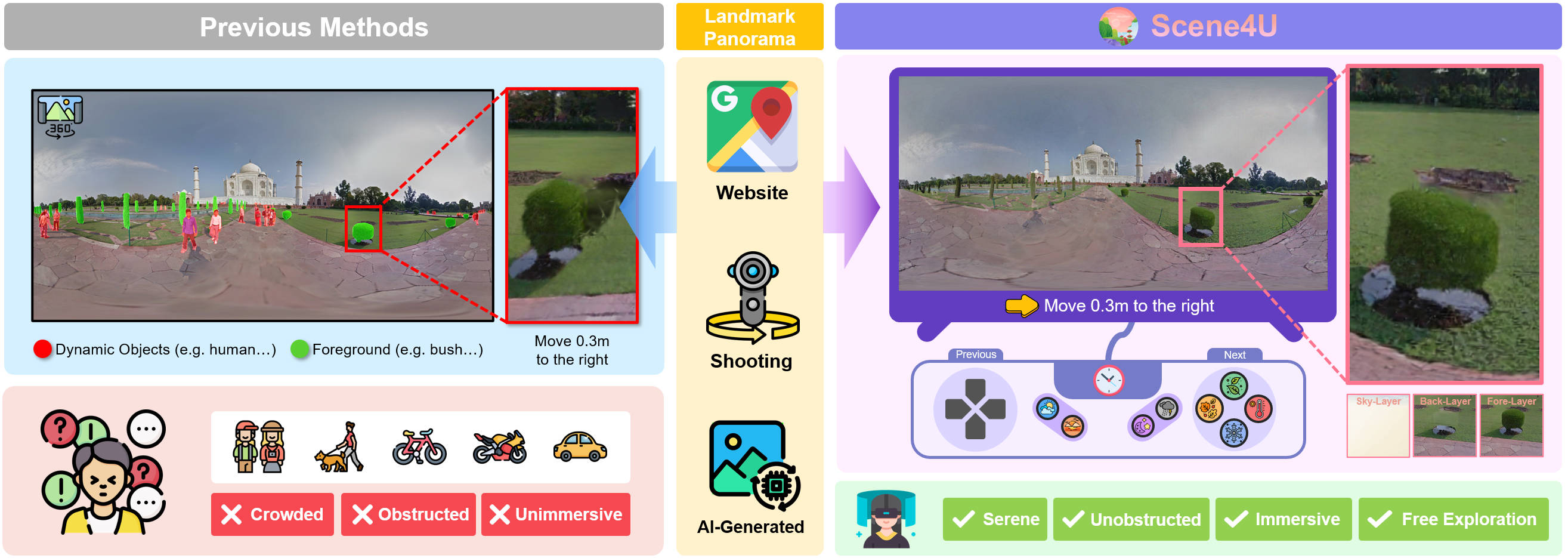}
\captionof{figure}{\textbf{Overview of Scene4U.} Scene4U is an unobstructed 3D scene construction framework based on single-view panoramas. By inputting a real panoramic image, Scene4U reconstructs a 3D scene free from dynamic objects such as pedestrians and vehicles, supporting unrestricted navigation.}
\label{fig:Scene4U}
\end{center}%
}]

\begin{abstract}
The reconstruction of immersive and realistic 3D scenes holds significant practical importance in various fields of computer vision and computer graphics. Typically, immersive and realistic scenes should be free from obstructions by dynamic objects, maintain global texture consistency, and allow for unrestricted exploration. The current mainstream methods for image-driven scene construction involves iteratively refining the initial image using a moving virtual camera to generate the scene. However, previous methods struggle with visual discontinuities due to global texture inconsistencies under varying camera poses, and they frequently exhibit scene voids caused by foreground-background occlusions. To this end, we propose a novel layered 3D scene reconstruction framework from panoramic image, named Scene4U. Specifically, Scene4U integrates an open-vocabulary segmentation model with a large language model to decompose a real panorama into multiple layers. Then, we employs a layered repair module based on diffusion model to restore occluded regions using visual cues and depth information, generating a hierarchical representation of the scene. The multi-layer panorama is then initialized as a 3D Gaussian Splatting representation, followed by layered optimization, which ultimately produces an immersive 3D scene with semantic and structural consistency that supports free exploration. Scene4U outperforms state-of-the-art method, improving by 24.24\% in LPIPS and 24.40\% in BRISQUE, while also achieving the fastest training speed. Additionally, to demonstrate the robustness of Scene4U and allow users to experience immersive scenes from various landmarks, we build WorldVista3D dataset for 3D scene reconstruction, which contains panoramic images of globally renowned sites. The implementation code and dataset will be made publicly available.
\end{abstract}    
\vspace{-15pt}
\section{Introduction}
\label{sec:intro}
The rapid development of virtual reality technology has opened up new possibilities to create immersive and realistic experiences. With virtual reality headsets, users can enjoy the breathtaking landscapes and unique cultures of various regions of remote travel without leaving the comfort of their home. However, high-quality immersive experiences rely on 3D generated scenes with high realism and consistency, which remains a significant challenge in artificial intelligence and 3D computer vision.

High-fidelity 3D scenes reconstruction utilize 3D reconstruction techniques, including traditional handcrafted \cite{bernardini1997sampling, bernardini1999ball, petitjean2001regular, kada20093d, wang2023building3d} and learning-based reconstruction methods \cite{yao2019recurrent, mildenhall2021nerf, kerbl20233d, lee2024compact}. The traditional methods relying on scanning technology achieve high-accuracy geometric structures and capture objects and spatial relationships in fine detail. However, they come at the cost of significant time and labor expenses. Moreover, scan-based reconstruction methods often fall short in texture quality, with the generated textures frequently lacking detail and realism. With the rapid advancement of deep learning, many new methods have been introduced for 3D reconstruction with photogrammetry \cite{ma2022multiview, zhang2023geomvsnet, wang2024dust3r, leroy2024grounding}, Neural Radiance Fields (NeRF) \cite{wei2021nerfingmvs, meng2023neat, wang2021neus, martin2021nerf}, and 3D Gaussian Splatting (3DGS) \cite{guedon2024sugar, lin2024vastgaussian, chen2025mvsplat, liu2025citygaussian}. However, these methods heavily depend on multi-view information, which is difficult to obtain in many practical scenarios, thereby limiting their applicability.

To improve the accessibility of multi-view data, several diffusion-based studies have been proposed that iteratively generate images from novel viewpoints for single-view 3D scene reconstruction \cite{hollein2023text2room, fridman2024scenescape, yu2024wonderjourney}, partially addressing the limitations of viewpoint availability. However, since only local texture information from existing images is used during the iterative process, the imperfect 3D scene lacks global consistency. This leads to significant visual discontinuities between different regions of the scene, which severely affects both immersion and realism.

Panoramic images, compared to perspective images, offer a broader coverage and richer contextual information, which result in better visual consistency and help address the visual discontinuities. Consequently, some researchers have begun exploring the extensive scene cues in panoramic images to improve the performance of generative models in 3D reconstruction \cite{gu2022omni, kulkarni2023360fusionnerf, wang2024perf}. We find that high fidelity and availability 3D scenes generation requires the integrity of the scene. However, due to significant foreground-background occlusion, the 3D space generated from panoramic images exhibits noticeable gaps and voids, which negatively affect the immersive visual experience.

Recent works such as WonderWorld \cite{yu2024wonderworld} and LayerPano3D \cite{yang2024layerpano3d} have highlighted the potential of layered representations for enhancing scene understanding. Building upon these insights, we propose a novel framework named \textbf{Scene4U}, which leverages real-world panoramic images to perform multi-layer 3D scene reconstruction, producing highly realistic and coherent immersive experience scenes.

(1) Following the input of prompt text and the original panoramic image, we first employ a Climate Controller to generate a spatiotemporally specific panoramic image. Subsequently, the generated panoramic image is processed using the open-vocabulary Semantic Segment Anything (SSA) \cite{chen2023semantic} instance segmentation model. Both the segmentation masks and the panoramic image are then fed into a large language model (LLM) for semantic filtering to obtain hierarchical masks.

(2) To elevate the panoramic image to a 360-degree scene representation, we perform layered inpainting based on the multi-layer masks, and employ depth estimation and completion methods to obtain multi-layer repaired scene, ultimately converting the panoramic image representation into a point cloud.

(3) The point cloud from the previous stage is initialized as a 3DGS representation, which is then optimized using a layered training strategy to address scene occlusion issues, resulting in a highly realistic and spatially consistent 3D scene.

Through the multi-stage hierarchical 3D generation, Scene4U enables the conversion of real panoramic images into a high-precision 3D scene, significantly improving the visual quality and consistency of the scene. It provides a more feasible strategy for virtual reality and immersive experience applications. Our main contributions are summarized as follows:

\begin{itemize}
\item We propose Scene4U, a layered scene reconstruction framework that generates highly consistent 3D scenes from a single real-world panoramic image. Scene4U effectively addresses challenges such as texture inconsistency and occlusions within complex environments, enabling users to immerse themselves in realistic scene exploration.

\item We develop a method that combines instance segmentation with a LLM to achieve effective foreground-background recognition. By leveraging the visual comprehension capabilities of the LLM, the proposed method improves the accuracy of classifying foreground and background regions, thereby facilitating multi-layered 3D scene reconstruction.

\item We provide a dataset for panoramic image 3D reconstruction, covering numerous famous landmarks worldwide. The dataset offers users a diverse array of real-world scenes, allowing them to freely explore renowned global attractions from the comfort of their own home in a virtual environment.

\end{itemize}
\section{Related Work}

\begin{figure*}[!ht]
    \centering
    \includegraphics[width=0.85\linewidth]{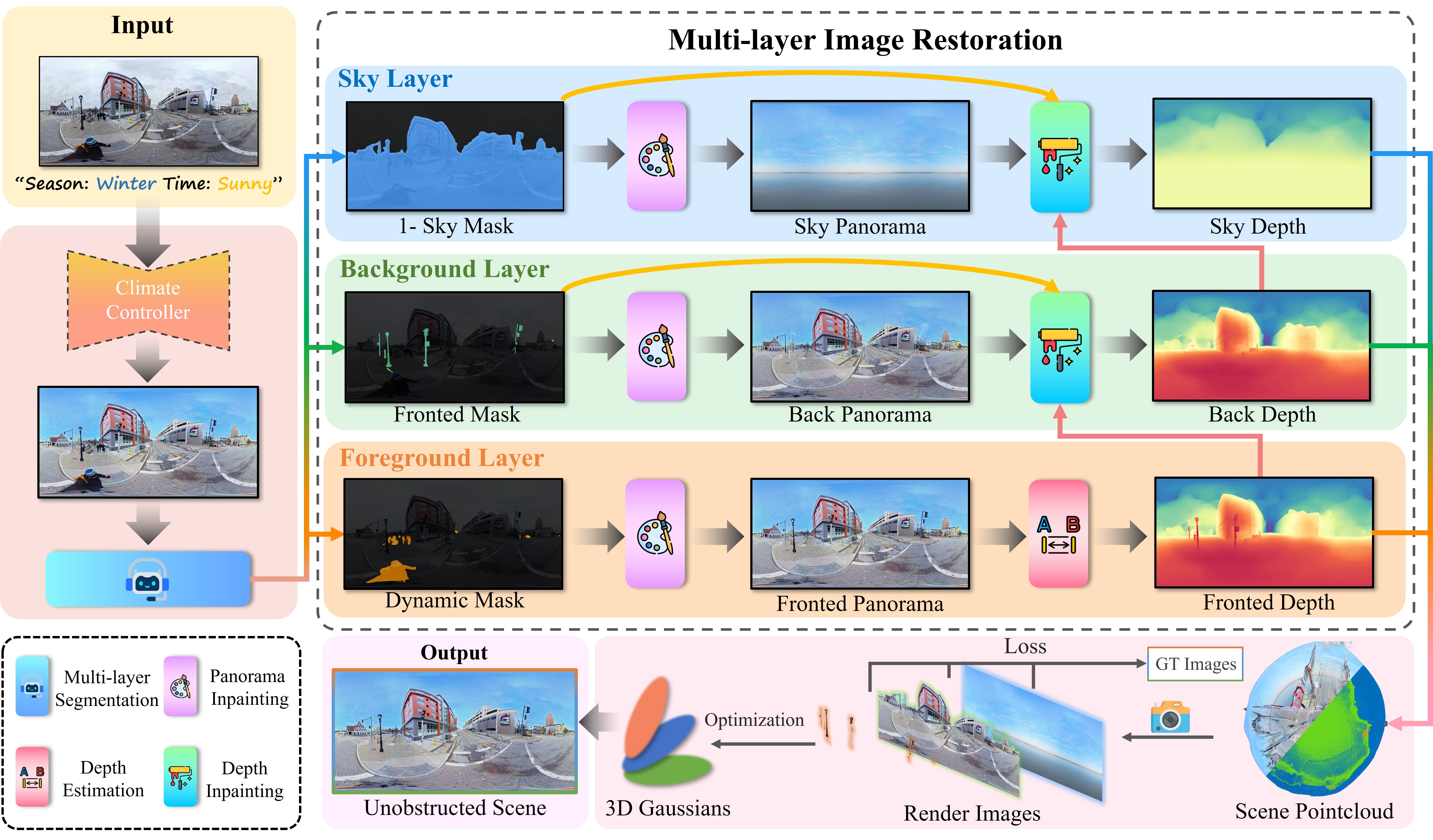}
    \caption{\textbf{The overview of Scene4U pipeline.} In the first stage, we use the input panoramic image and text prompts to generate a panoramic image with corresponding spatiotemporal characteristics through Climate Controller, followed by multi-layer segmentation. In the second stage, we use the obtained multi-layer mask results to perform multi-layer construction on the panoramic scene image. In the third stage, we apply a layered training strategy to optimize the scene, reconstructing an immersive environment for free exploration.}
    \vspace{-12pt}
    \label{pipeline}
\end{figure*}

\subsection{Image Variation Based on Diffusion Models}
Image transformation refers to generating diverse style variations based on a given image sample, while preserving the original semantic information and basic visual perception of the image. In recent years, with the substantial advancements in Diffusion Models in the field of image generation, diffusion model-based image transformation methods have been widely applied to tasks such as style transfer \cite{zhang2023inversion, wang2023stylediffusion}, novel view synthesis \cite{liu2023zero, liu2024one, chan2023generative}, and image editing \cite{kawar2023imagic, zhang2023sine, mou2024diffeditor}. Limited by the long intervals between data collections, street view data from the same location often presents a monotonous scene environment, which brings challenges to the construction of multi-styled realistic scenes. To address the above challenges, we employed a text-guided image editing approach to temporally initialize the original input scene images, generating the target street-view panoramic images as required by users, with specific details provided in Section 3.1.

\vspace{-5pt}

\subsection{3D Scene Representation}
Traditional 3D scene representations utilize point cloud \cite{berger2014state}, volume \cite{lombardi2019neural, nguyen2019hologan}, and meshes \cite{zhang2001efficient}. While each of these methods has its own advantages, they typically require large amounts of data, leading to high computational costs. Moreover, these methods struggle to meet the high-quality rendering requirements necessary for immersive scenes. With the rapid development of deep learning technologies, implicit representation methods, such as NeRF \cite{mildenhall2021nerf}, have demonstrated outstanding capabilities in novel view synthesis and high-quality rendering. However, these methods still face challenges in terms of optimization efficiency and rendering speed. To overcome these limitations, subsequent research has introduced explicit representation techniques on top of implicit methods to enable faster training and rendering \cite{yu2021plenoxels, xu2022point}. Among these, the 3DGS \cite{kerbl20233d} approach, which is based on Gaussian kernels, achieves real-time rendering and exceptional rendering quality through the use of alpha blending and differentiable rasterization techniques. Therefore, we adopt 3DGS as our method for scene representation in this work.

\begin{figure*}[ht]
    \centering
    \begin{subfigure}[t]{0.47\linewidth}
        \centering
        \includegraphics[width=\linewidth]{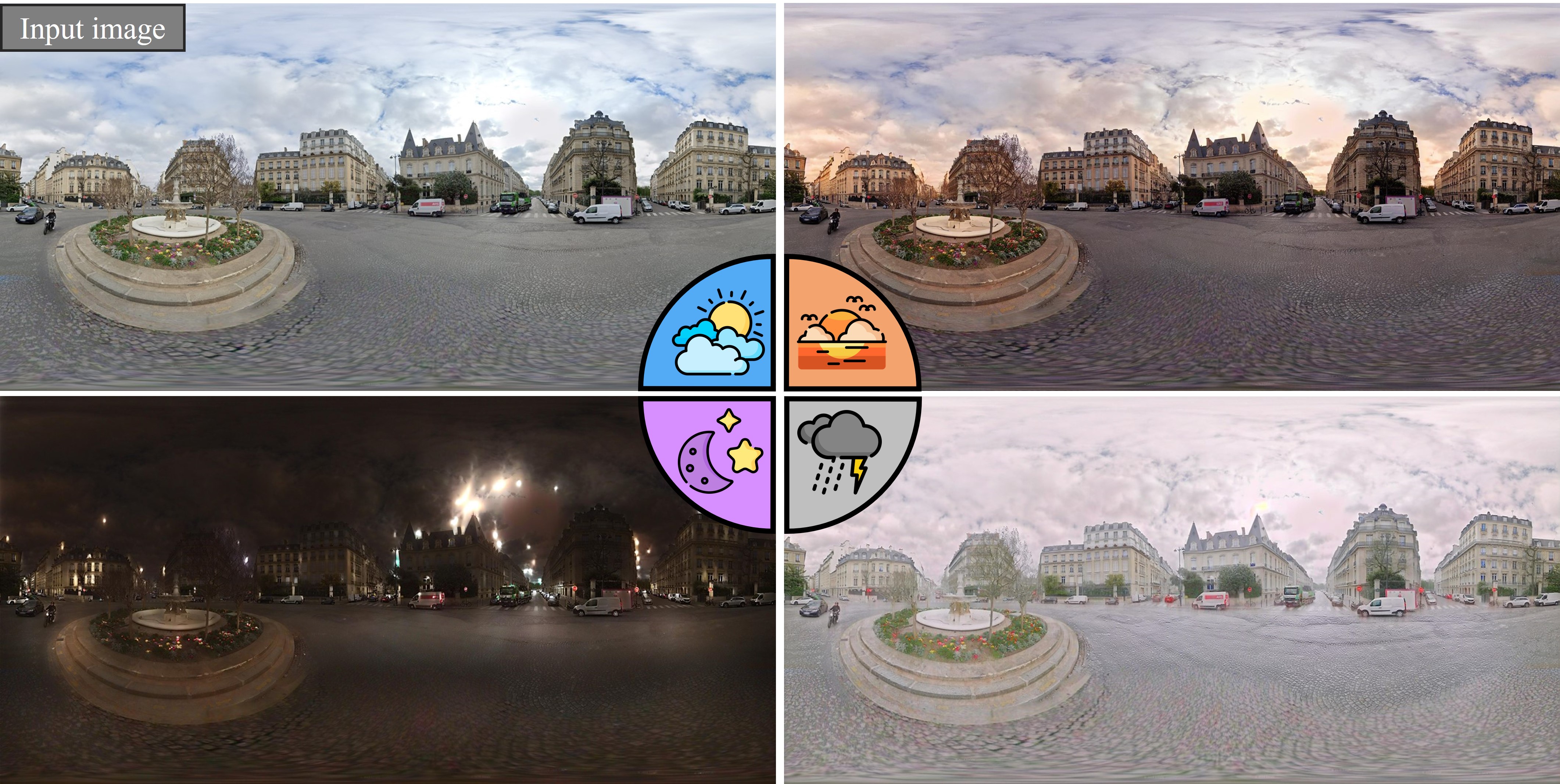}
        \caption{Time of day}
        \label{img2img_left}
    \end{subfigure}
    \hfill
    \begin{subfigure}[t]{0.47\linewidth}
        \centering
        \includegraphics[width=\linewidth]{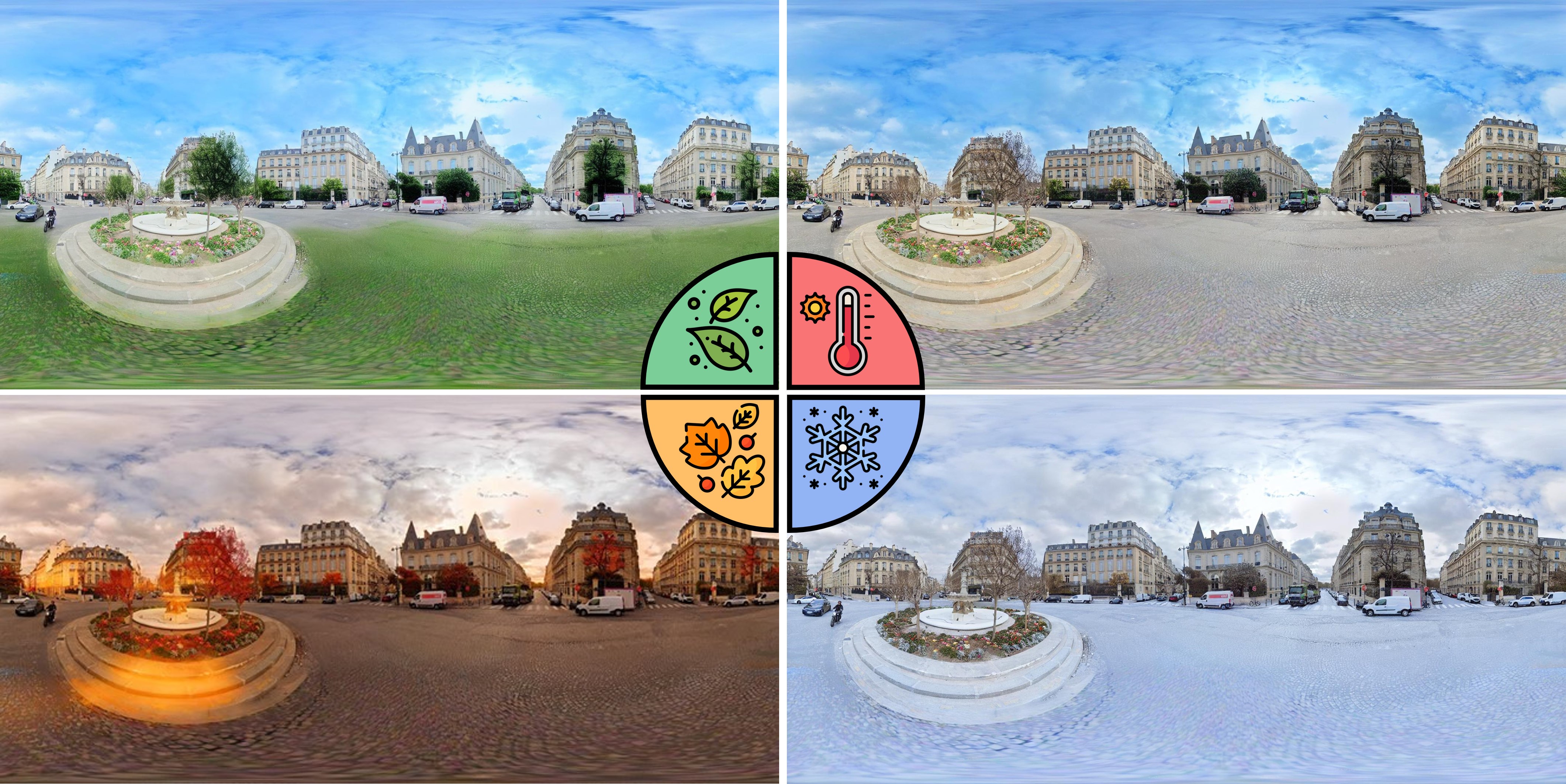}
        \caption{Four seasons of the year}
        \label{img2img_right}
    \end{subfigure}
    \vspace{-5pt}
    \caption{Illustration of Climate Controller synthesis results. The Climate Controller module can generate realistic street-view images under various weather and time conditions, enhancing the diversity of reconstructed scenes.}
    \vspace{-14pt}
    \label{fig:img2img}
\end{figure*}

\vspace{-5pt}

\subsection{3D Scene Generation}
Currently, most of the existing methods depend on multi-view images for 3D scene generation, which infer the 3D structure of a scene from images captured from various perspectives \cite{li2024dngaussian, chen2024mvsplat, lin2024vastgaussian, zhou2024drivinggaussian}. However, these methods cannot be directly applied to single-view scene generation. In the field of single-view scene generation, existing methods such as LucidDreamer \cite{chung2023luciddreamer}, RealmDreamer \cite{shriram2024realmdreamer} and Text2Immersion \cite{ouyang2023text2immersion} iteratively refine the scene by moving a virtual camera, gradually generating a complete 3D scene by capturing and inpainting information from different angles. However, during the iterative refinement process, these methods only utilize partial texture information of the scene, neglecting the global consistency, leading to obvious visual discontinuities in the generated 3D scene. In DreamScene360 \cite{zhou2025dreamscene360}, researchers used panoramic images to build 3D scenes, thereby ensuring the consistency of overall scene information. However, due to insufficient consideration of foreground-background occlusion relationships, DreamScene360 can only allow scene viewing from fixed perspectives and does not support free navigation. 

WonderWorld \cite{yu2024wonderworld} and LayerPano3D \cite{yang2024layerpano3d} have pioneered the concept of distinguishing the spatial relationships among objects in a scene through scene layering.

Inspired by these methods, we employ a semantic-aware layered construction and rendering approach, which removes the restrictions of fixed viewpoints and allows users to navigate within the scene, providing a more immersive experience.

\section{Scene4U}

We propose a panoramic image-driven framework for immersive 3D scene reconstruction that removes distracting elements (e.g., pedestrians and vehicles) to render a 3D scene with high visual consistency and scene integrity. The key insight is allowing users to explore scenes from any time. As shown in Fig.~\ref{pipeline}, Scene4U consists of three main stages. First, a text-driven diffusion model generates a target panoramic image with specific spatiotemporal properties. Next, a large language model assists in decomposing the target panoramic image into multiple layers. These decomposed layers are then processed with image inpainting and depth restoration to obtain complete layered scene information. Finally, we transform the multi-layered panoramic images into a multi-layered 3D scene by 3DGS refinement, creating an immersive scene.

\vspace{-2pt}
\subsection{Climate Controller}

To address the limited scene environmental conditions in real-world panoromatic images, we first introduce a text-guided Climate Controller based on Instruct Pix2Pix \cite{brooks2023instructpix2pix}. The climate controller applies environmental condition constraints to the input panoramic images, generating corresponding images in the target domain by specifying conditions such as season, time of day (e.g., daytime or nighttime), etc. These generated images are then used as input for the subsequent layered panoramic reconstruction. As shown in Fig.~\ref{fig:img2img}, the Climate Controller enables the synthesis of diverse scene environments, producing images with various times and weather conditions.

\subsection{Layered Panorama Construction}

Complex dynamic foreground objects in real-world scenes cause occlusion, making parts of the background invisible. To this end, we propose a novel layered representation method for 3D reconstruction, which eliminates the invisible background and constructs a complete 3D scene by dividing it into multiple visual layers.

\begin{figure}[!t]
    \centering
    \setlength{\abovecaptionskip}{0.1cm}
    \includegraphics[width=1\linewidth]{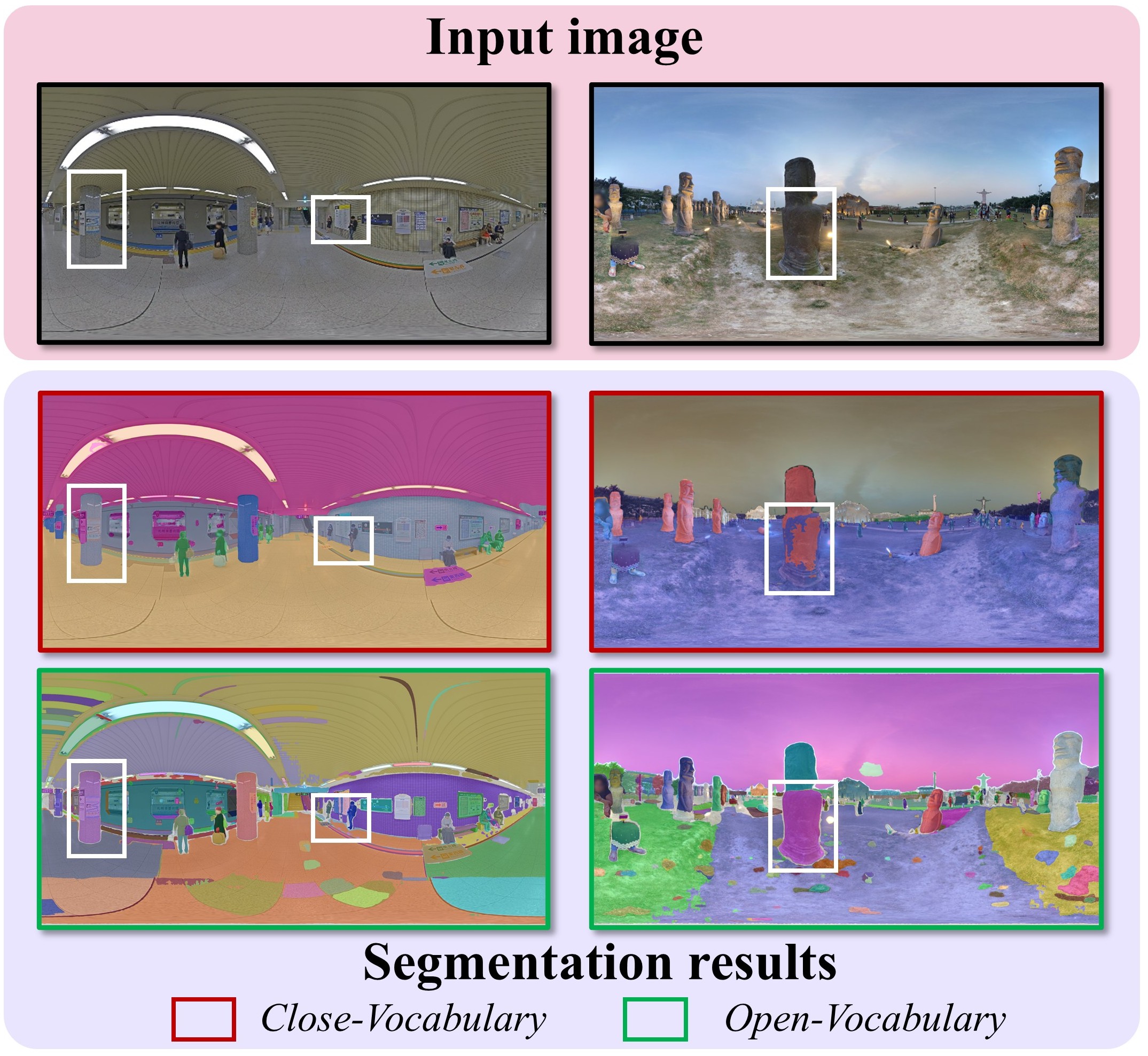}
    \caption{The segmentation results comparison of the close-vocabulary and open-vocabulary segmentation.}
    \vspace{-5pt}
    \label{fig:seg_case}
\end{figure}

\noindent\textbf{Semantic-Aware Layer Parsing.} Multi-layer representation requires accurate identification of hierarchical objects in the scene.
LayerPano3D\cite{yang2024layerpano3d} innovatively proposes scene layering method using instance segmentation prior. Instance segmentation methods are generally categorized into close-vocabulary \cite{jain2023oneformer} and open-vocabulary \cite{chen2023semantic} approaches. As illustrated in Fig.~\ref{fig:seg_case}, the former is often constrained by a predefined set of categories, which limits its adaptability to the complexity of real-world scenarios.
To address this, we propose a novel approach that effectively resolves the applicability issues of existing methods in complex environments.

\begin{figure}[!t]
    \centering
    \includegraphics[width=1.0\linewidth]{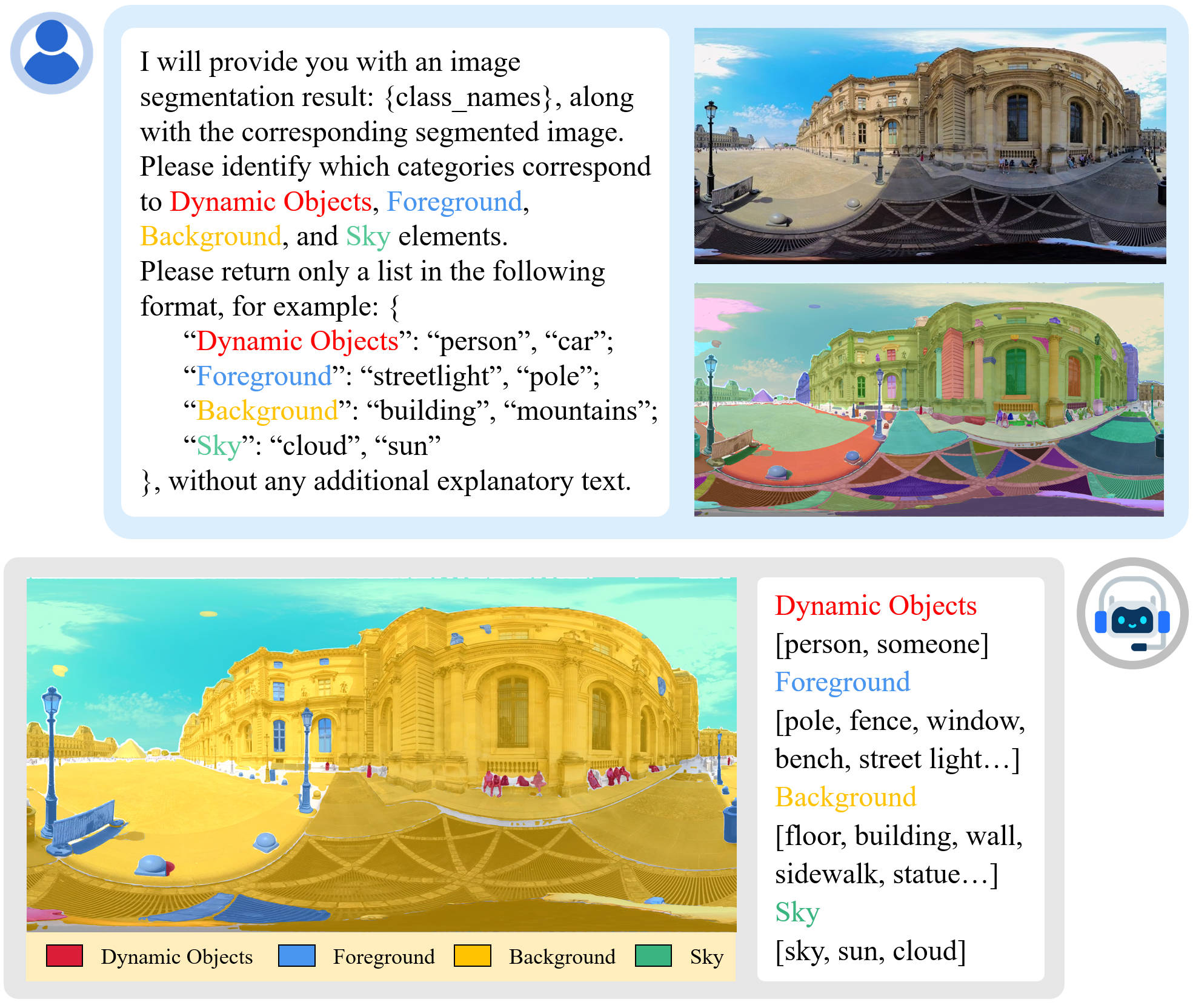}
    \caption{Illustration of multi-layer segmentation strategy. Starting with initial open-vocabulary segmentation labels for the panorama images, we utilize LLM to group categories and output masks for dynamic objects, foreground, background, and sky regions, respectively.}
    \vspace{-6pt}
    \label{seg}
\end{figure}

Firstly, we predefine hierarchical category representations of the scene, including sky $L_{sky}$, background $L_{background}$, foreground $L_{foreground}$ and dynamic object $L_{dynamic}$ layer, which are sequentially numbered as 3, 2, 1, and 0, respectively. Secondly, we employ a fine-grained open-vocabulary instance segmentation \cite{chen2023semantic} to extract individual semantic objects from the panoramic image. We observe that the predicted instances are difficult to directly categorize into hierarchical layers based on semantic information, especially for class-agnostic masks. Therefore, we utilize LLM \cite{qwen}  to interactively perform semantic filtering and hierarchical classification of objects. Specifically, we input the original RGB image and corresponding segmentation masks into the LLM and use predefined hierarchical categories for prompting. The LLM then outputs the hierarchically filtered segmentation masks. Finally, we obtain multi-layer segmentation masks with spatial hierarchy, as shown in Fig.~\ref{seg}. Notably, we define objects that are likely disappear in the short term as the dynamic objects. Removing dynamic objects helps generate immersive scenes that are more spacious and natural. While there are challenges in the definition of categories, the segmentation results (Fig.~\ref{seg}) are robust to occlusion issues and geometric orderings.

\setlength{\belowcaptionskip}{-0.2cm}

\setlength{\belowcaptionskip}{-0.2cm}

\noindent\textbf{Occlusion-Guided Layer Recovery.} The occlusion presents an inevitable integrity issue in layered image for 3D scene construction. To address this, we use the FLUX-inpainting to progressively repair each layer using the hierarchical masks. Based on the given mask (e.g., the areas to be restored), we utilize image context information to render and fill in the matching pixels. Layer-by-layer repair ensures semantic consistency across layers while restoring occluded content, resulting in high-quality generated scene. For instance, the foreground mask is used to help the repair of occluded areas in the background, while the background mask, together with the foreground mask, assists in repairing of occluded sky regions. During the repair process, we use 'no objects present' as a prompt to guide the repair model, allowing it to seamlessly transition and restore the original details of the scene when filling in the content of each layer. This process is formalized as:
\begin{equation}
    \tilde{I}_{l+1} = Inpaint_{RGB}(\tilde{I}_{l}, M_{l+1}),
\end{equation}
where $Inpaint_{RGB}$ is the function of FLUX-inpainting, $\tilde{I}_{l}$ and $M_{l+1}$ denote the RGB panoramic image of the $l$-{th} layer and the mask for the $(l+1)$-th layer, respectively. The repaired image, obtained after filling the missing regions, maintains both structural and textural consistency as well as completeness. 

\noindent\textbf{Hierarchical Spatial Inference.} To construct a complete and spatially accurate 3D scene, we perform depth estimation and depth completion on the layered images to ensure spatial consistency across them. Inspired by LayerPano3D \cite{yang2024layerpano3d}, we utilize 360MonoDepth \cite{rey2022360monodepth} to project the panorama image onto 20 overlapping perspective patches and use a pretrained monocular depth model ZoeDepth \cite{bhat2023zoedepth} to calculate depth for each projection. To address the affine ambiguities in scale and displacement, we align each projection by inputting the optimized parameters into a multi-scale and spatially varying deformation field, to get a high-resolution panoramic depth map. We use the existing depth information and RGB texture to predict and complete the masked regions without depth value, mathematically expressed as follows:
\begin{equation}
    \tilde{D}_{l} = Inpaint_{depth}(\tilde{I}_{l}, D_{l+1}, M_{l}),
    \label{depth-completion}
\end{equation}
where $Inpaint_{depth}$ denotes the model for depth completion. $D_{l+1}$ represents the depth completion result for the $(l+1)$-th layer, $\tilde{I}_l$ is the RGB panorama of the $l$-th layer, and $M_{l}$ denotes the mask of the $l$-th layer, respectively. Notably, to ensure that the spatial relationship between the depth information of the background and sky layers remains consistent with that of the foreground layer, we perform depth estimation exclusively for the foreground layer, while using depth completion \cite{liu2024infusion} for the background and sky layers. 

\subsection{Multi-layer Panoramas to 3D Scene}

\textbf{Panorama to Point Cloud.} The restored panoramic image is converted into a point cloud to construct a 2D-3D representation. Using the equidistant projections of the sphere, we efficiently convert the 2D panorama into 3D point cloud without additional computational overhead. Specifically, for each pixel located at $ (i, j) $ in a panoramic image $ \tilde{I} \in \mathbb{R}^{H \times W} $, we calculate its latitude angle $ \theta $ and longitude angle $ \phi $ using Eq.~\ref{pixel2angle}. Then, we derive the corresponding 3D coordinates $ (X, Y, Z) $ for each pixel $ (i, j) $ based on its latitude, longitude, and depth values, as Eq.~\ref{pixel2point}.

\begin{equation}
    \theta_{i,j} = \frac{\pi i}{H}, \quad \phi_{i,j} = -\frac{2\pi j}{W} + \pi,
    \label{pixel2angle}
\end{equation}
\begin{equation}
    \begin{pmatrix} 
        X \\ Y \\ Z   
    \end{pmatrix} = d_{(i,j)} \times 
    \begin{pmatrix}
        cos\phi_{i,j} \cdot cos\theta_{i,j} \\ cos\theta_{i,j} \\ sin\theta_{i,j} \cdot cos\phi_{i,j}
    \end{pmatrix}
    \label{pixel2point}
    % (X, Y, Z) = d_{(i,j)} * (cos\phi_j cos\theta_i, sin\phi_j, cos\phi_j sin\theta_i)
\end{equation}
Following on the above formula, we sequentially calculate the 3D coordinates of each pixel in the panorama, forming a point cloud that integrates both visual texture and spatial structure. To ensure the accuracy of the multi-layer structure and the independence of point cloud, we only convert the pixels within the mask after filtering pixels in each layer. Finally, We initialize the multi-layer point cloud as 3DGS strictly following the layer index.

\noindent\textbf{Layered Numbered Optimization.} 3DGS is able to accurately present the 3D representation of a scene captured from multi-view images. Unsatisfactorily, rendered 3DGS panorama results in distortion due to the significant differences in projection form and viewpoint characteristics between panoramic and perspective views. Unlike previous methods, we decompose and transform the panorama into standard perspective views from different angles through camera field-of-view observations. We set up a set of cameras to cover the panoramic area for viewpoint sampling, where each camera shares the same intrinsic matrix but has its own independent extrinsic matrix. This setup ensures that the camera group observes different regions of the 3D space with consistent viewpoint parameters. Following Eq.~\ref{E2P}, we capture the perspective views after sampling the panorama and use them as the ground truth for that viewpoint:
\begin{equation}
\begin{aligned}
    x_e &= \frac{x \cdot FOV_x + 2\theta_0 + 2\pi}{4\pi} \cdot W_e, \\
    y_e &= \frac{y \cdot FOV_y + 2\phi_0 + \pi}{4\pi} \cdot H_e 
    \label{E2P}
\end{aligned} 
\end{equation}
where \( x_e \) and \( y_e \) represent the pixel coordinates in the panorama, \( x \) and \( y \) are the normalized coordinates in the perspective image, \( FOV_x \) and \( FOV_y \) denote the horizontal and vertical field of view, \( \theta_0 \) and \( \phi_0 \) are the center angles for longitude and latitude, and \( W_e \) and \( H_e \) indicate the width and height of the panorama, respectively.

We find that occlusion effects lead to inconsistencies in the layered scene. In response to the above difficulties, LayerPano3D \cite{yang2024layerpano3d} took the lead in proposing a solution by optimizing the 3DGS model layer by layer in multi-layer panoramas using foreground asset masks to achieve scene refinement. Followed by LayerPano3D \cite{yang2024layerpano3d}, we perform independent optimization for each layer during the 3DGS refinement. Due to the clear hierarchical structure of the scene, where the back layers do not occlude the front layers, starting the optimization from the back layers helps to establish foundational background information and ensures depth consistency across layers.Therefore, We extract the 3DGS of specific layers in depth order (from the back layers to the front) for optimization. In the training phase, we set the loss function as a weighted sum of $ \mathcal{L}_{1} $ and $ \mathcal{L}_{D-SSIM} $:
\vspace{-2pt}
\begin{equation}
\begin{split}
    \mathcal{L} = &(1 - \lambda)\mathcal{L}_{1}(I_{render}^{l}, I_{gt}^{l} \odot M_{l}) + \\
    &\lambda \mathcal{L}_{D-SSIM}(I_{render}^{l}, I_{gt}^{l} \odot M_{l}),
\end{split}
\label{loss-cal}
\end{equation}
where the hyperparameter $ \lambda $ is set to 0.2. $ \odot $ denotes the intersection between ground truth image and layered mask, to prevent the pixel value from other regions from interfering with the loss function. Moreover, the optimization process is performed separately for each layer, masking the loss function gradients of other layers to prevent affecting the parameters of those layers.

\begin{figure*}[!t]
    \centering
    \includegraphics[width=0.98\linewidth]{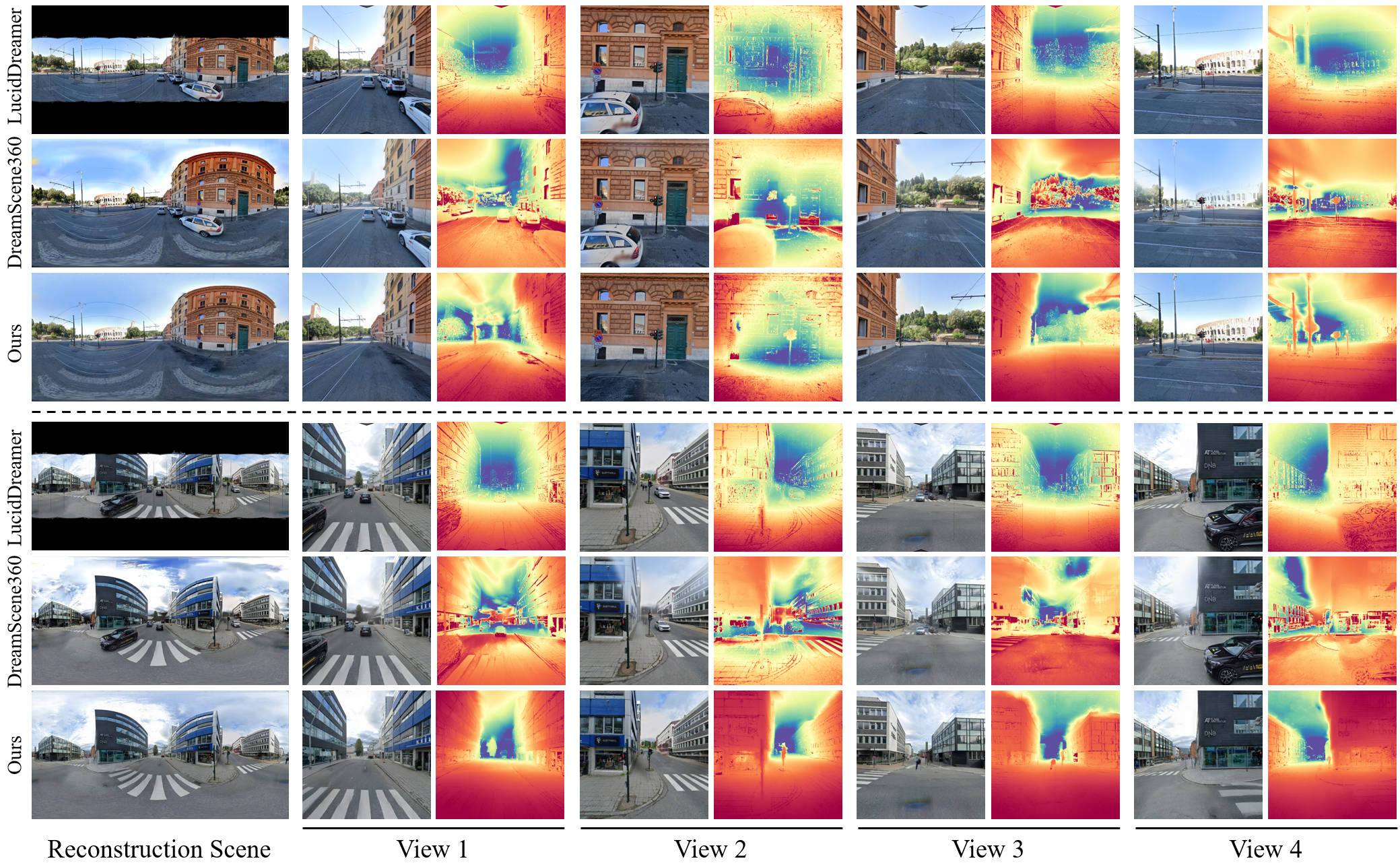}
    \vspace{-2pt}
    \caption{Qualitative comparison of scene reconstruction results from different methods, including LucidDreamer, DreamScene360, and our Scene4U. Our method generates open scenes without any dynamic object occlusions. Benefiting from the layered construction strategy, Scene4U produces scenes with richer hierarchical structures while maintaining overall consistency.}
    \label{result}
\end{figure*}
\section{Experiments}
\begin{table*}[!t]
    \centering
    \begin{tabular}{l|ccc|cc|c}
    \toprule
    Method & PSNR ($\uparrow$) & SSIM ($\uparrow$) & LPIPS ($\downarrow$) & NIQE ($\downarrow$) & BRISQUE ($\downarrow$) & Training Time ($\downarrow$) \\
    \midrule
    LucidDreamer \cite{chung2023luciddreamer} & 30.409 & \textbf{0.985} & 0.033 & 5.299 & 50.513 & 12 min 37 s\\
    DreamScene360 \cite{zhou2025dreamscene360} & 29.546 & 0.922 & 0.047 & 4.445 & 34.574 & 18 min 23 s \\
    \midrule
    \rowcolor{gray!12}
    \textbf{Ours(w/ Dynamic Layer)} & \underline{31.237} & 0.931 & \underline{0.030} & \underline{3.793} & \underline{28.892} & \underline{11 min 33 s} \\
    \rowcolor{gray!24}
    \textbf{Ours} & \textbf{32.778} & \underline{0.959} & \textbf{0.025} & \textbf{3.605} & \textbf{27.793} & \textbf{11 min 13 s} \\
    \bottomrule
    \end{tabular}
    \caption{Qualitative comparison of scene reconstruction results of different methods on WorldVista3D. Scene4U outperforms DreamScene360, improving by 24.24\% in LPIPS and 24.40\% in BRISQUE, and achieves the fastest training speed. The best results are in \textbf{bold}.}
    \vspace{-5pt}
    \label{tab:comparison}
\end{table*}

\subsection{WorldVista3D Dataset}

To provide a diverse immersive experience of real-world scenes and to verify the robustness of the Scene4U, we build a 3D real-world scene reconstruction dataset of famous landmarks--\textbf{WorldVista3D}, with 120 panoramic images of well-known tourist attractions worldwide, obtained from the Google Street View API 
. All image resolutions in WorldVista3D are resampled to a uniform scale of $ 2,048 \times 1,024 $. Note that this resolution is also used as the input scale for training our model of Layered Panorama Reconstruction.

\subsection{Implementations Details}

In the Layered Panorama Reconstruction stage, we first resample the input  panoramic images to a resolution of $ 2,048 \times 1,024 $. Then, we apply the Semantic Segment Anything model to generate initial instance segmentation masks. Using QWen-Plus-Latest, we differentiate segmentation labels and produce corresponding masks for the multi-layer panorama. The FLUX.1 model is  used for image inpainting to obtain repaired panoramic results for each layer. Finally, the 360monodepth and depth-inpainting models are employed for depth estimation and restoration.

At the Multi-layer Panorama-to-3D Scene stage, we generate an initial scene point cloud along the camera’s ray direction based on the depth map, and assign a 3DGS index to each layer for subsequent layered optimization. During the 3DGS refinement stage, we independently train the Gaussian Spheres for each layer according to the assigned indices, utilizing supervised perspective images at a resolution of $ 512 \times 512 $. For the training setup of the Gaussian points in each layer, we set the number of iterations to $3,000$, $4,000$, and $3,000$ for the sky, background, and foreground layers, respectively, while disabling the splitting and pruning operations of the 3DGS. All experiments are conducted on an NVIDIA A100 80G GPU.

\begin{figure*}[!ht]
    \centering
    \includegraphics[width=0.93\linewidth]{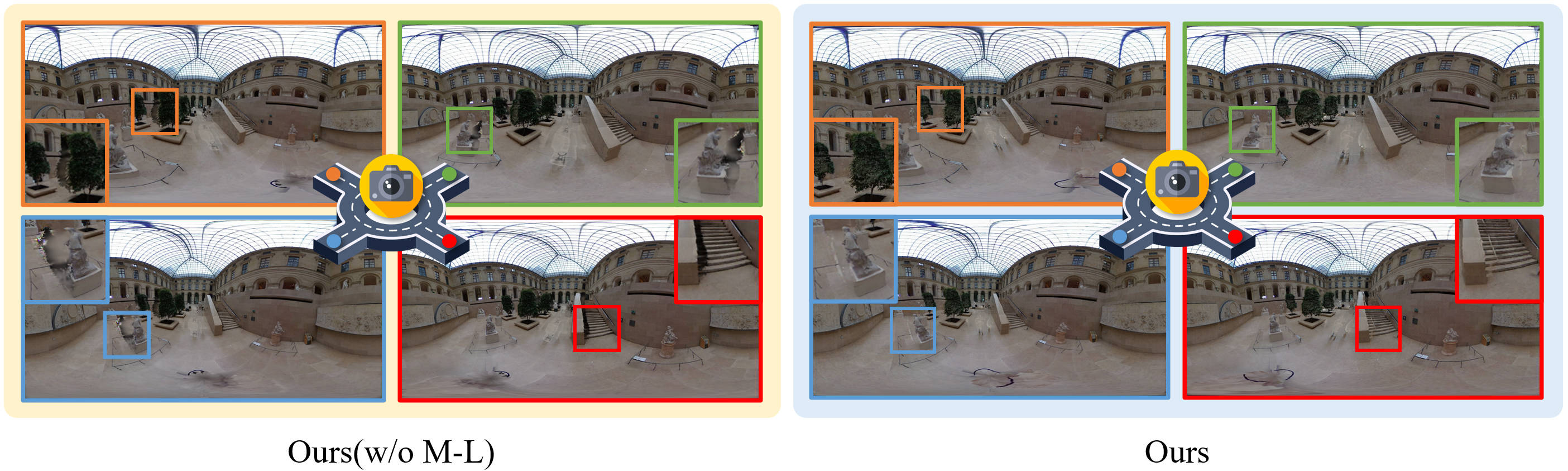}
    \caption{Qualitative comparison of results using the Layered Panorama Reconstruction strategy. M-L indicates the use of the Layered Panorama Construction strategy, with orange / green / blue / red boxes representing rendering results after moving the same distance of 20\% to the front-left, front-right, back-left, and back-right from the scene center, respectively.}
    \vspace{-8pt}
    \label{ablation}
\end{figure*}

\subsection{Evaluation Metrics}

To comprehensively evaluate the performance of Scene4U in reconstruction quality and rendering robustness, we employ classic reconstruction metrics alongside no-reference image quality assessment metrics. Specifically, to assess the reconstruction quality of the scene, we select the Structural Similarity Index Measure (SSIM), Peak Signal-to-Noise Ratio (PSNR), and Learned Perceptual Image Patch Similarity (LPIPS) \cite{zhang2018unreasonable} to calculate the similarity between rendered image and reference image. Additionally, to evaluate the robustness of the rendering results, we use traditional image quality assessment metrics, including the Natural Image Quality Evaluator (NIQE) \cite{mittal2012making} and the Blind / Referenceless Image Spatial Quality Evaluator (BRISQUE) \cite{mittal2012no}, to assess the quality of no-reference images.

\subsection{Comparisons with Other Methods}

We select publicly available methods, LucidDreamer \cite{chung2023luciddreamer} and DreamScene360 \cite{zhou2025dreamscene360}, as baselines for performance comparison with our proposed Scene4U. LucidDreamer takes a single image and textual prompts as input, employing a progressive inpainting strategy to generate a 360-degree scene. This method incrementally fills and expands the image content through multiple iterations to create a complete view. Since the original LucidDreamer does not support panoramic image input, we modify it to accept an input panorama as the initial scene to ensure a fair evaluation of the baseline methods. In contrast, DreamScene360 generates a 360-degree panoramic 3DGS scene based on text prompts, converting textual descriptions into a complete and high-quality 3D scene. To ensure a fair comparison, we excluded the Climate Controller module from the comparative experiments and additionally computed the metrics without removing dynamic objects.

Fig.~\ref{result} presents the qualitative comparison results of different methods. Our method achieves the best performance in five metrics compared to previous state-of-the-art methods. Scene4U surpasses DreamScene360 in scene details, achieving finer reconstruction results. As shown in Tab.~\ref{tab:comparison}, Scene4U achieves improvements of 7.79\% and 24.24\% in PSNR and LPIPS, respectively. The quantitative results demonstrate the stability and robustness of the proposed method in panoramic image reconstruction. Moreover, the proposed method demonstrates optimal training efficiency, proving the lightweight nature of layered reconstruction.
\begin{table}[!t]
    \centering
    \setlength{\abovecaptionskip}{0.02cm}
    \setlength{\tabcolsep}{1mm}
    \begin{tabular}{c|c|cc}
    \toprule
     Move Distance & Method & NIQE ($\downarrow$) & BRISQUE ($\downarrow$) \\
    \midrule
    \multirow{2}{*}{0.1 m} & Ours(w/o M-L)  & 3.015 & 31.721 \\
    & Ours & \textbf{2.861} & \textbf{31.539} \\
    \midrule
    \multirow{2}{*}{0.2 m} & Ours(w/o M-L)  & 3.053 & 33.367 \\
    & Ours & \textbf{2.930} & \textbf{32.038} \\
    \midrule
    \multirow{2}{*}{0.3 m} & Ours(w/o M-L)  & 3.323 & 36.224 \\
    & Ours & \textbf{3.126} & \textbf{35.011} \\
    \bottomrule
    \end{tabular}
    \caption{Quantitative comparison of multi-layer scene reconstruction under different movement scales. M-L denotes the Layered Panorama Construction strategy. The best results are in \textbf{bold}.}
    \vspace{-10pt}
    \label{tab:ablation}
\end{table}
\setlength{\belowcaptionskip}{-0.5cm}

\subsection{Ablation Study}

The ablation study on the Layered Panorama Reconstruction is presented in Fig.~\ref{ablation}. Compared to the single-layer panorama reconstruction, the generated scene reveals background areas that are occluded by the foreground objects from new perspectives when the user moves within the scene. Overall, layered reconstruction effectively addresses the voids caused by foreground-background occlusions. The quantitative analysis results of different configurations under continuous movement in a specified direction within the scene are presented in Tab.~\ref{tab:ablation}. The results demonstrate the proposed Layered Panorama Reconstruction strategy improves robustness over longer distances, achieving 5.93\% and 3.35\% improvements in NIQE and BRISQUE against the single-layer panorama scene after moving 0.3 m, respectively. These results confirm that our generated realistic 3D scenes meet user demands for unrestricted navigation.
\section{Conclusion}

In this work, we propose a novel and effective framework, named Scene4U, for generating highly realistic and consistent 3D scenes through a hierarchical reconstruction strategy. First, we stylize the input panorama and perform multi-layer decomposition with the assistance of an open-vocabulary segmentor and LLMs. Subsequently, we remove dynamic and foreground objects from the scene and design layered inpainting of the occluded areas based on image texture and depth information. Finally, the multi-layer panorama is initialized as a 3DGS representation and hierarchically refined to construct immersive scenes with semantic and structural consistency. Comprehensive experimental results demonstrate that Scene4U outperforms previous methods in terms of visual quality and achieves a realistic visual experience, especially after the removal of dynamic object layers. Scene4U is also competitive in efficiency. Overall, our Scene4U provides a novel solution for reconstructing multi-temporal, dynamic-object-free, globally consistent, and explorable immersive 3D scenes from panoramic images.
{
    \small
    \bibliographystyle{ieeenat_fullname}
    \bibliography{main}
}
\end{document}